\documentclass[preprint,12pt]{elsarticle}

\usepackage{setspace}
\usepackage[numbers]{natbib}
\usepackage{amsmath,amsfonts}
\usepackage{color}
\usepackage{amssymb}
\usepackage{amsmath}

\usepackage{tabularx}
\usepackage{booktabs}
\usepackage{siunitx}
\newcolumntype{Y}{>{\centering\arraybackslash}X} 
\usepackage{threeparttable}
\usepackage{multirow}
\usepackage{amssymb}
\usepackage{array}
\usepackage{paralist}       
\usepackage{amsmath} 
\usepackage{pifont}

\usepackage{stfloats}
\usepackage{subcaption}

\def\vs{\emph{v.s.}}

\journal{Elsevier}

\begin{document}
\definecolor{gray}{rgb}{0.5, 0.5, 0.5}  
\begin{frontmatter}

\title{Towards Subject-Oriented Video Captioning via User-Specified Targets}

\author[1]{Yunchuan Ma}
\ead{mayunchuan23@mails.ucas.ac.cn}
\author[1]{Chang Teng}
\author[1]{Guorong Li}
\author[2]{Yuankai Qi}
\author[1]{Laiyun Qing}
\author[1]{Qingming Huang}



\address[1]{University of Chinese Academy of Science, Beijing,100190, China}
\address[2]{Macquarie University}


\begin{abstract}
Describing video content according to users' needs is a long-held goal.
Although existing video captioning methods have made significant progress,
the generated captions may not focus on the entity that users are particularly interested in.
To address this problem, we propose a new video captioning task, \textbf{S}ubject-\textbf{O}riented \textbf{V}ideo \textbf{C}aptioning (SOVC), which aims to allow users to specify the describing target via a bounding box.
To support this task, we construct two subject-oriented
video captioning datasets based on two widely used video
captioning datasets: MSVD and MSRVTT, by annotating
subjects in each video for each caption. 
These datasets pave the way for describing users' interested targets.
%
To tackle this task, we introduce a method tailored to this task, named SOVCNet.
It consists of two key components: a subject-oriented sampling module that samples frames related to the subject to minimize irrelevant information; and a subject-oriented encoding module that utilizes the subject areas as hard prompts and integrates learnable soft prompts, enhancing the model's focus on the subject's activities and facilitating adaptation to the downstream generation task.
Extensive experimental results demonstrate the effectiveness of our method on this new task.
\end{abstract}

\end{frontmatter}



\begin{figure}[!t]
\centering
\includegraphics[width=0.8\linewidth]{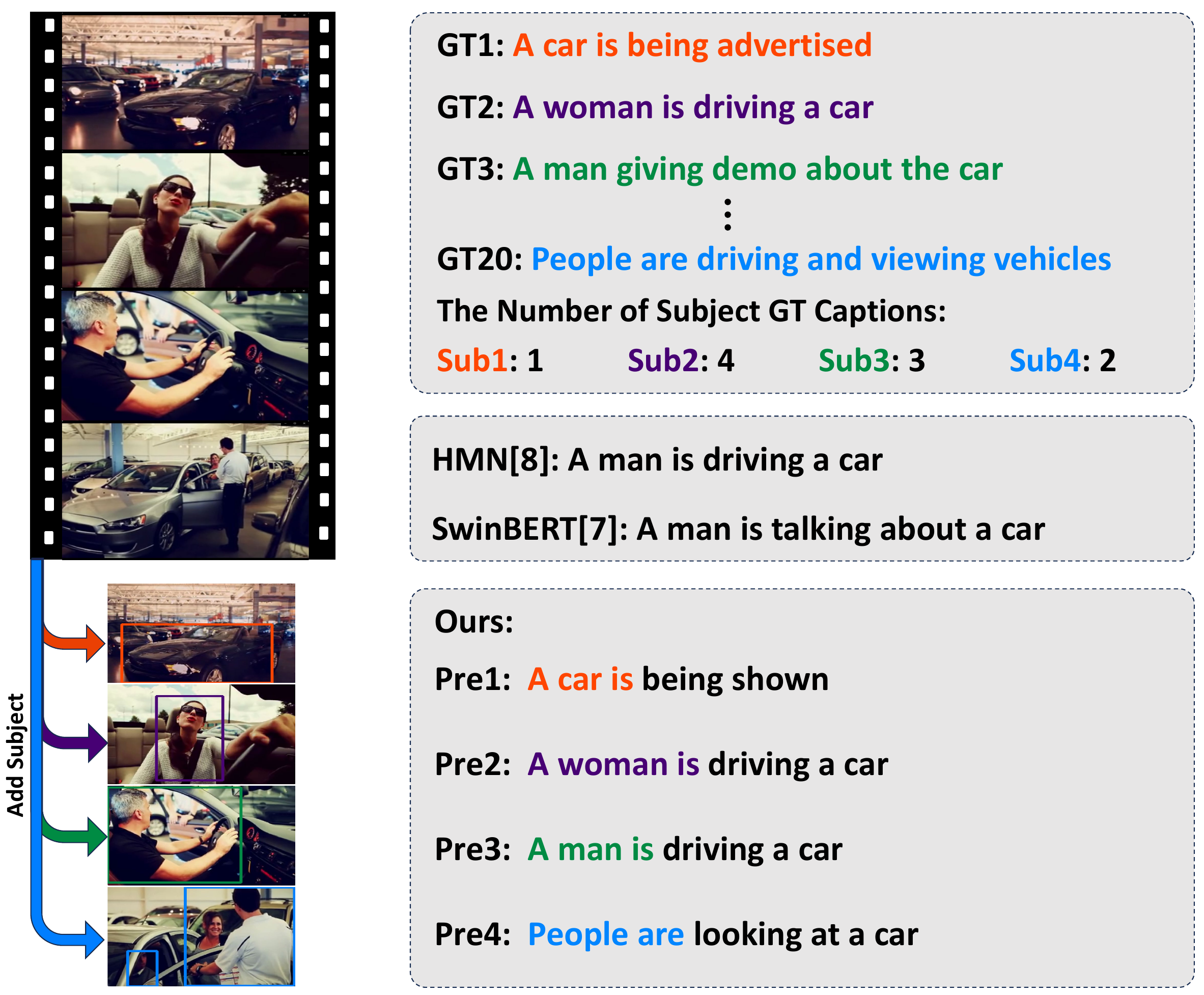}
\caption{Conventional video captioning (upper panel) \vs our subject-oriented video captioning (bottom panel), where by introducing an additional user-specified subject, the model is guided to generate video descriptions focused on the subject.
}
\label{fig:insight}
\end{figure}

\section{Introduction}

{V}{ideo} captioning~\cite{savc, retta} is an important video understanding task~\cite{multi-task-ptal, TRA}, aiming to understand and accurately describe the video content in a single sentence.
It plays a crucial role in numerous applications, such as video title generation, blind assistance, and video search.
Unlike static images~\cite{spl1, tmm1, rs1}, videos contain additional temporal information, making video understanding more challenging.
However, existing methods 
do not allow users to specify the target of interest and therefore the generated captions might not be the desired.
This could limit its application in real-world scenarios. 
Consider a video with multiple individuals, as shown in Figure~\ref{fig:insight}. If a user
wants a summary of a specific individual’s behaviors, such as the woman, traditional video captioning methods often fail. 
The state-of-the-art methods SwinBERT~\cite{swinbert} and HMN~\cite{hmn} generate ``A man is talking about a car''   and ``A man is driving a car'', respectively. This is because they are trained to cover the majority of entities mentioned in the ground truth, which is ``a man'' in this case.
Consequently, they cannot cater to user-specific needs for targeted descriptions.

In the field of image captioning, similar attempts have been made. For instance, a framework~\cite{cornia2019show} named “Show, Control and Tell" generates captions by taking a series of image regions as input. However, this method requires users to select and rank multiple entity areas, which is less practical in the video scenario because it is unreasonable to require the user to watch the whole video first and then pick out interacted objects. A desired method should just require to specify the target of interest and let the model its own to discover what happens to the target.
Another limitation of applying image captioning methods for video captioning is their inability to handle temporal information in multiple frames, such as complex actions. 
The above-mentioned limitations also apply to other image captioning methods, such as CGO~\cite{cgo} and VSR-guided model~\cite{vsr}.

Another line of related research is controlled video captioning. Existing methods primarily focus on controlling sentence structure, as demonstrated by \cite{wang2019pos+cg}, which introduces Part-of-Speech (POS) information for this purpose. 
SCVC~\cite{saat} aims to generate one caption that describes the video contents semantically while also imitating the syntactic structure of the given example sentence. However, these methods lack effective control over the subjects in the videos. Meanwhile, although O2NA~\cite{o2na} allows for control using simple entity-class words, it lacks the ability to select a specific entity if multiple instances of the same class exist, thus lacking specification flexibility and precision.

To address the aforementioned problems, we propose a new task named ``subject-oriented video captioning'', which aims to generate descriptions centered on activities of the user-specified entity. 
As an example, in the bottom panel of Figure~\ref{fig:insight},  when specifying the car as the description target, the generated caption is ``A car is being shown''; when the target of interest is changed to the woman, the generated caption changes to ``A woman is driving a car''. 
However, compared to conventional video captioning, the new task faces two main challenges: 1) how to capture the most relevant information related to the subject in the video. 2) how to model the relationship between the subject and video information.
%

To tackle these challenges, we propose a subject-oriented video captioning Network (SOVCNet), which incorporates two of our proposed modules (subject-oriented sampling module and subject-oriented encoding module) into existing methods. 
Conventional video captioning methods uniformly sample fixed video frames to perceive the overall video content.
In our task, we argue the effectiveness of traditional sampling methods because our objective is to describe the subject's activities rather than the overall video content.
Thus, we develop a new subject-oriented sampling method specifically designed for our proposed task.
We first use an image encoder to extract features of the subject and frames and compute their cosine similarity.
Then, based on the cosine similarity, we use a cluster-based method to sample a fixed number of frames.
After sampling subject-related video frames, we further consider how to leverage these visual information to model the relationship between the subject and other entities in the video. 
Inspired by the success of P-Tuning~\cite{liu2021gpt} and Visual Prompt Tuning (VPT)~\cite{jia2022visual}, we design a new subject-oriented encoding module.
This module utilizes subject features as hard prompts to direct the video encoder's attention toward the subject's activities. Additionally, it plugs learnable soft prompts that enable the model to adapt to the generation task.

To support our proposed task, we re-annotate two widely used video captioning datasets: MSVD and MSR-VTT.
Specifically, we identify frames that match the video description and label the coordinates of the subjects.
%
%
More details about the annotations and dataset are provided in Section~\ref{sec:dataset}.

Our main contributions can be summarized as:
\begin{enumerate}
  \item We propose a novel task: subject-oriented video captioning, which allows the users to focus on objects of interest, especially in some real-world application scenarios, such as surveillance videos, etc.
  \item We propose a new model for subject-oriented video captioning, which contains two modules: the subject-oriented sampling module and the subject-oriented encoding module. The former samples frames related to the subject to reduce unnecessary information. The latter uses the subject area as hard prompts and plugs the soft prompts to make the model focus more on the subject's activities and adapt to the generation task.
  \item We constructed two new datasets named SO-MSVD and SO-MSRVTT specifically for the subject-oriented video captioning problem. 
  \item Extensive experiments on the newly built datasets demonstrate our model outperforms various baselines for the subject-oriented video captioning task.
\end{enumerate}

\section{Related work}
In this section, we briefly review general video captioning,  controllable video captioning and prompt learning, which are related to our task and put this work in a proper context.

\noindent\textbf{General Video Captioning}
Recent research on video captioning primarily focuses on the design of network architectures.
Early methods apply simple CNN architectures for representation learning from videos, such as S2VT~\cite{venugopalan2015s2vt}. 
Later, more complex encoder architectures are proposed.
STG-KD~\cite{stg-kd} and ORG-TRL~\cite{org-trl} develop encoders using object relation graphs to distinguish the spatial-temporal relationships among video objects.
SGN~\cite{sgn} proposes a semantic grouping network, which encodes videos into semantic clusters by identifying relevant frames.
HMN~\cite{hmn} adopts a hierarchical modular design to encode three levels of visual features to get comprehensive video representations.
GL-RG~\cite{DBLP:conf/ijcai/gl-rg} creates a  global-local encoder, which exploits rich temporal representation for video captioning.
Meanwhile, some work also focuses on improving the decoding ability, 
such as Yap~\textit{et al.}~\cite{/iccv/YaoTCBPLC15} integrating temporal attention mechanisms and Pei~\textit{et al.}~\cite{marn} incorporating attended memory decoders.
Beyond obtaining richer visual representations and enhancing decoding capabilities, the long-tailed problem remains a challenge, as low-frequency tokens are crucial for generating detailed content. RSFD~\cite{rsfd} addresses this by persistently capturing the linguistic representations of these rare tokens, while TextKG~\cite{textkg} integrates an additional knowledge graph to mitigate the challenge of long-tail words.

Different from the above methods, SwinBERT~\cite{swinbert} employs an end-to-end transformer-based design,
which directly takes a sequence of video frames as inputs and uses caption-related loss to adjust feature learning.
However, it increases the computational burden and time consuming to handle the dense video tokens.
To improve inference efficiency, Cocap~\cite{cocap} directly uses compressed video as input, enabling faster caption generation. 
Most recently, some work~\cite{LAVENDER,MV-GPT} leverages large pre-train models as the backbone to align powerful vision and language representations.

Although conventional video captioning methods, as mentioned above, are capable of generating accurate overall descriptions of a video clip, they fall short when it comes to emphasizing a particular entity of interest as the subject.
%
In this work, we pioneer the task of subject-oriented video captioning, shifting the focus towards the activities of a specific subject within a video, rather than pursuing depicting general information in the video. 
Moreover, to accommodate our novel task, we have restructured the traditional captioning datasets and explore several attempts based on state-of-the-art general video captioning methods.

\noindent\textbf{Controllable Video Captioning}
In contrast to general video captioning, much less attention is paid to controllable video captioning.
Some efforts involve using additional sentences as reference signals, allowing the generated sentences to learn the syntactic structure or style of the reference sentences.
For example,
SMCG~\cite{smcg} models the syntactic structure from a given exemplar sentence and designs a modified LSTM module to transfer the syntax in the exemplar to new captions for the input video. 
FS-StyleCap~\cite{Stylized_vc} extract the style of stylized example to generate sentences with the same style.
SAVC~\cite{savc} proposes a two-stage framework to learn seven different styles and generate more specific captions.
On the other hand, some works employ special inputs as control signals.
MSG~\cite{MSG} uses a controllable masked scene graph to represent interested objects and develops a corresponding decoder to infer masked relationships between objects to obtain the final caption.
In~\cite{o2na}, a method to generate captions about the specified objects is developed. 
To this end, object detection is performed, and then a non-autoregressive generator has to be used to make up a complete sentence from ahead-determined objects.
ERA~\cite{group_vc} introduces the reference videos to capture contextual information and further refine the group video captioning.   

Our task/method differs from the above-mentioned methods in two aspects: (I) The target selection is not limited to the nodes of a scene graph of MSG. 
Our task/method allows the user to select any entity in the video as the desired caption subject. Furthermore, the caption content is from the whole video, not limited to all the graph nodes of MSG.
(II) The additional control signals we utilized are unique, setting our approach apart from previous methods.
 
\noindent\textbf{Prompt Learning}
In recent years, prompt learning has made significant progress in the NLP field.
Prompts can be separated into hard prompts and soft prompts.
The former usually uses fixed text templates to guide the model in producing the desired responses, which are direct and clear.
GENPET~\cite{Schick} and PET~\cite{Schick2} employ pre-defined templates in a few-shot learning framework for text classification and conditional text generation tasks.
Wallace et al.~\cite{Wallace} applies a gradient-based search over actual tokens to automatically find the suitable templates described in a discrete space.
The latter employs learnable parameters to refine the model’s adaptation to specific downstream tasks.
Prefix-tuning~\cite{li2021prefix} delivers performance comparable to full fine-tuning while reducing parameter storage by a factor of 1000.
Additionally, prompt-relevant parameters can also be fine-tuned together with all or some of the parameters of the pre-trained models. PADA~\cite{ben2021pada} and P-Tuning~\cite{liu2021gpt} are notable examples. This method can better adapt to downstream tasks and achieve optimal performance.

Beyond the NLP domain, prompt learning also demonstrates its effectiveness in computer vision tasks.
For example, Visual Prompt Learning~\cite{jia2022visual} utilizes visual prompts to enhance pre-trained vision models, enabling specific tasks to be performed more effectively with less training data, which resonates with our hard-soft prompt fusion strategy.
Besides, ProTrack~\cite{yang2022prompting} facilitates multimodal tracking using non-learnable prompts.Meanwhile, the finding of Chen et al.~\cite{li2022grounded} — that spatially grounded prompts reduce model hallucination — provides theoretical support for our subject-crop hard prompts.

Inspired by the aforementioned work, we integrate subject information as hard prompts and learnable soft prompts to enhance the model's understanding of the user-specified subject and adapt to the downstream generation task. 
\section{Dataset}\label{sec:dataset}

To support our proposed subject-oriented video captioning task, we require a corresponding dataset that meets our research needs. Each video in the dataset includes $N$ chosen subjects, with corresponding activity descriptions for each subject.

\begin{table}[htbp]
    \caption{Statistics of our Subject-Oriented MSVD and  MSRVTT datasets.}
  \begin{center}
  \resizebox{ 0.7\linewidth}{!}
    {\begin{tabular}{c|cccccc} 
    \toprule
      \textbf{Dataset} &\textbf{Train}& \textbf{Val} & \textbf{Test}  & \textbf{Region} & \textbf{Frame} & \textbf{Caption} \\
      \midrule
      SO-MSVD & 1,871 & 150 & 1,287 & 4,287 & 3,411 &72,530\\
      SO-MSRVTT & 15,306 & 1,085 & 6,979  & 35,499 &  25,988 & 149527\\
      \bottomrule      
    \end{tabular}}
     \label{statistical information}
  \end{center}
\end{table}

\subsection{Dataset Construction}~\label{dataset construction}
We first construct two datasets based on two widely used general video captioning datasets: MSVD~\cite{msvd} and MSRVTT~\cite{msrvtt}.

\begin{figure*}[!t]
\centering
\includegraphics[width=\linewidth]{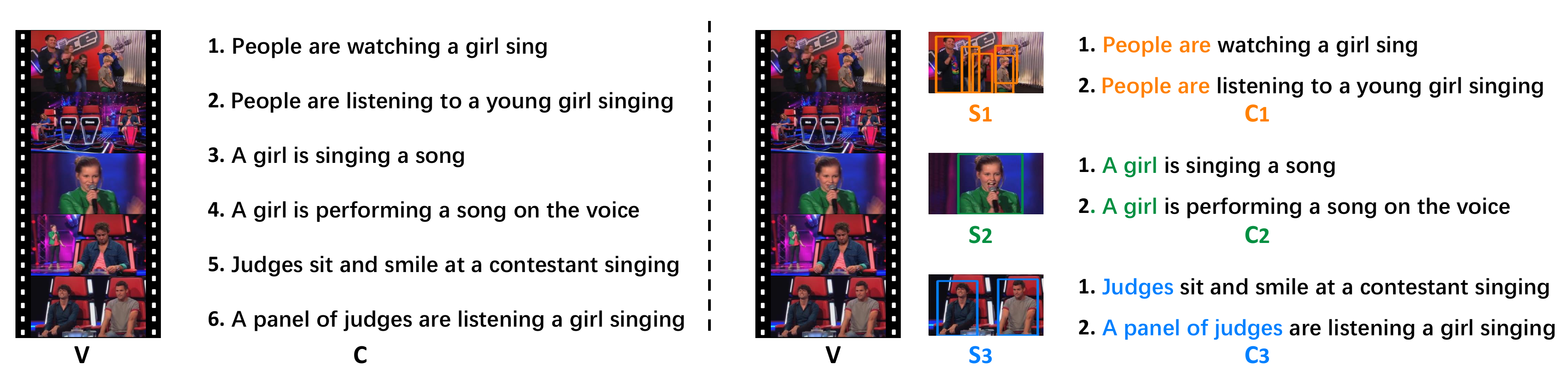}
\caption{Conventional video captioning dataset (left panel) \vs our Subject-oriented video captioning dataset (right panel).}
\label{fig:dataset}
\end{figure*}

We annotate the dataset in four steps.
First, we use \textit{spaCy}\footnote{https://spacy.io/} to extract entities in each caption.
%
We term these extracted entities as subjects.
Captions with abstract subjects like ``a video'' are discarded.  
Then, we annotate the bounding box in videos for each subject. We only annotate the frame where the subject appears for the first time.
To facilitate the annotation, we perform object detection using Faster-RCNN, then calculate and rank the cosine similarity between the entity categories predicted by Faster-RCNN and the subject words.
%
Next, we manually check each bounding box for each caption. Missing and incorrect bounding boxes are corrected during this step.
%
Finally, we reorganize the dataset as \(\{V, S, C\}\), where \(S\) represents the annotated subject regions within videos, \(V\) denotes the video, and \(C\) is the captions corresponding to the subject. 
As shown in right panel of Figure~\ref{fig:dataset}, one video can have multiple subjects, and each subject may have multiple captions. 

\subsection{Dataset Statistics}

Table~\ref{statistical information} presents the detailed specifics of the reannotated and reorganized datasets. In the subsequent sections, we elaborate on each of them, drawing upon the data shown in the table.

\begin{figure}[!ht]
\centering
\includegraphics[width=0.7\linewidth]{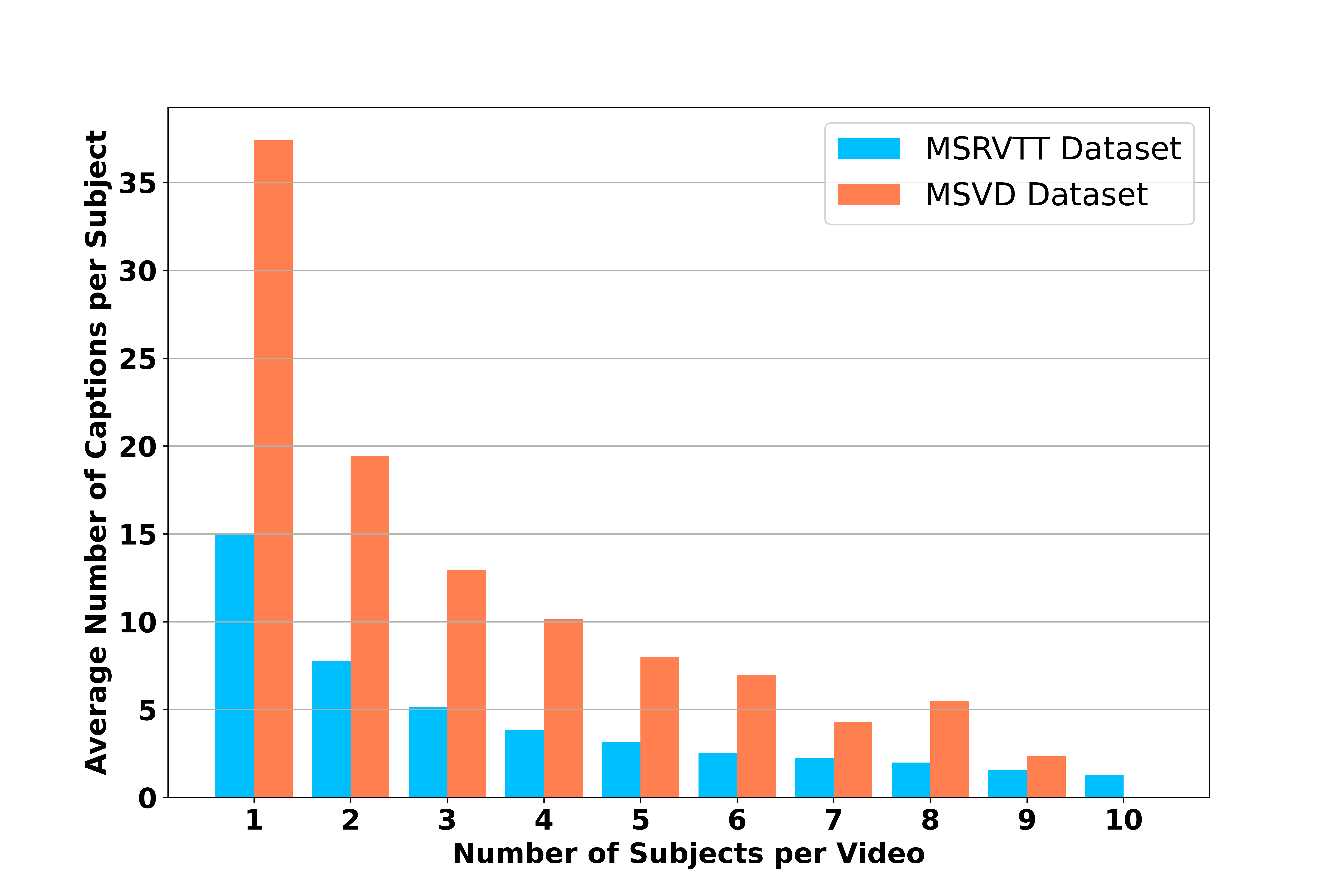}
\caption{ Average amount of captions regarding subject complexity.
}
\label{fig:average_captions_histogram}

\end{figure}

\begin{figure}[!ht]
\centering
\includegraphics[width=0.8\linewidth]{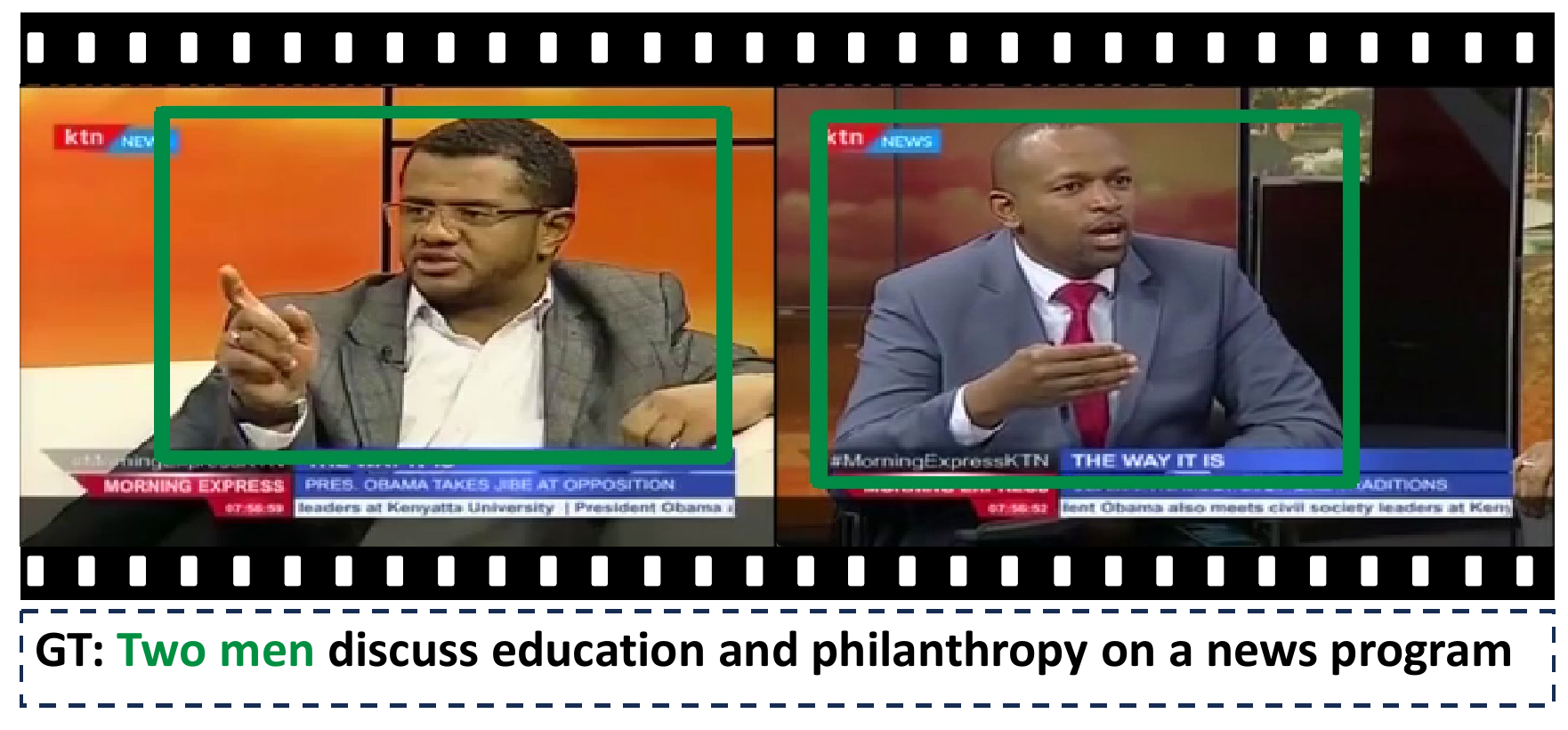}
\caption{ Multiple entities appear in different frames
}
\label{fig:different_frame}

\end{figure}

\noindent\textbf{Subject-Oriented MSVD} 
After filtering out captions with abstract subjects from the original MSVD dataset~\cite{msvd}, we annotate 3,308 subject samples for 1,961 videos, with 72,530 captions. On average, one subject corresponds to  22 sentences.
As shown in Figure~\ref{fig:average_captions_histogram}, subject-oriented MSVD contains a varying number of subjects ranging from 1 to 9, with the corresponding average number of captions approximately between 3 and 38.
Note that one subject may correspond to multiple frames and regions in a video. 
For example, consider the scenario where two men are talking about something as shown in Figure~\ref{fig:different_frame}; these `two men' may first appear in different frames, and each may occupy different regions even in the same frame. 
Based on our statistics, a total of 4,287 regions have been annotated across 3,411 frames. 
Since the total number of videos remains similar to its original scale, we adopt the same dataset division, leading to 1,871, 150, and 1,287 subject samples in the training, validation, and test splits, respectively.

\begin{figure}[t]
  \centering
  \subfloat[\footnotesize Subject-Oriented MSVD]{%
    \includegraphics[width=0.38\linewidth,trim=0 5pt 0 5pt,clip]{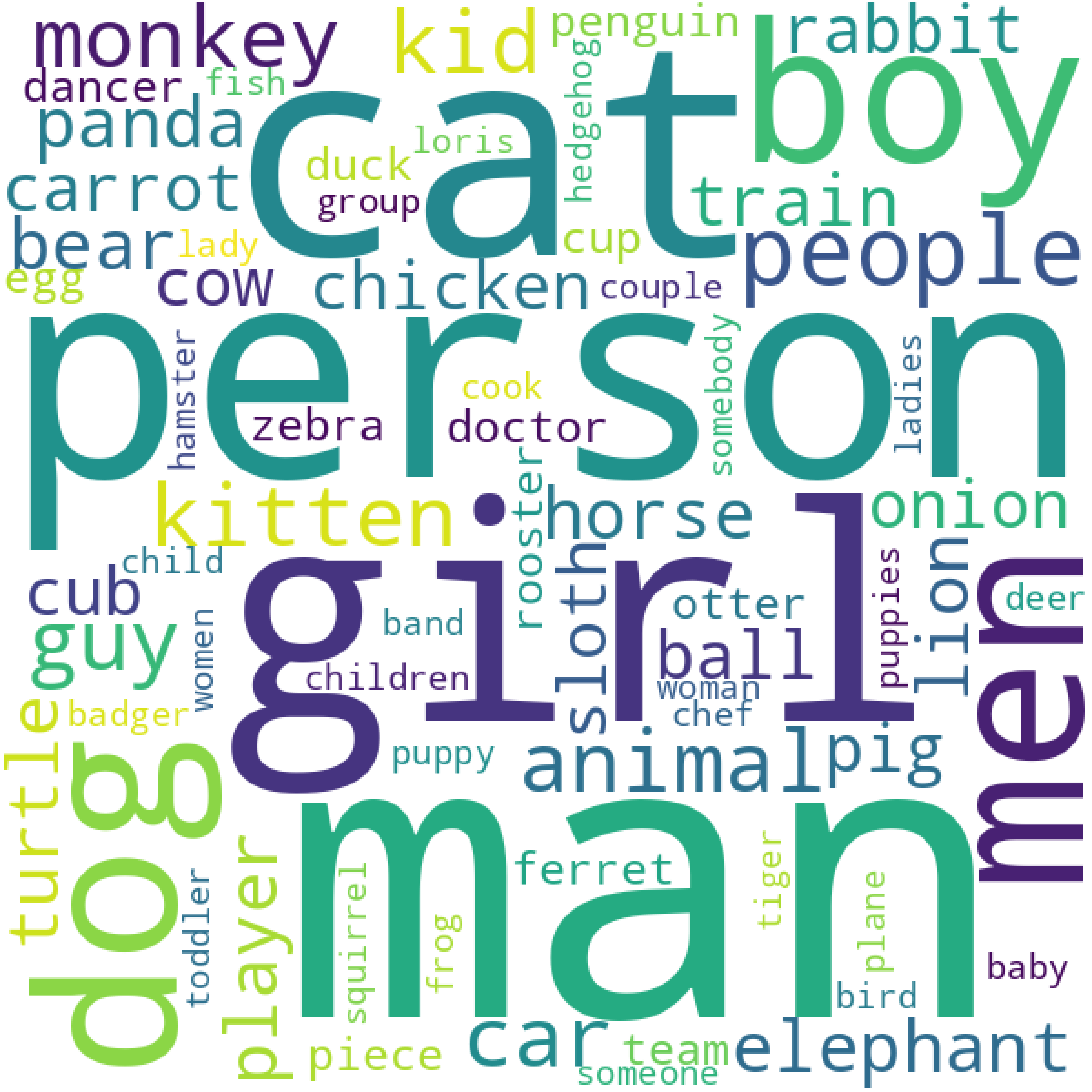}
    \label{fig:word_cloud_a}%
  }\hspace{0.04\linewidth} 
  \subfloat[\footnotesize Subject-Oriented MSRVTT]{%
    \includegraphics[width=0.38\linewidth,trim=0 5pt 0 5pt,clip]{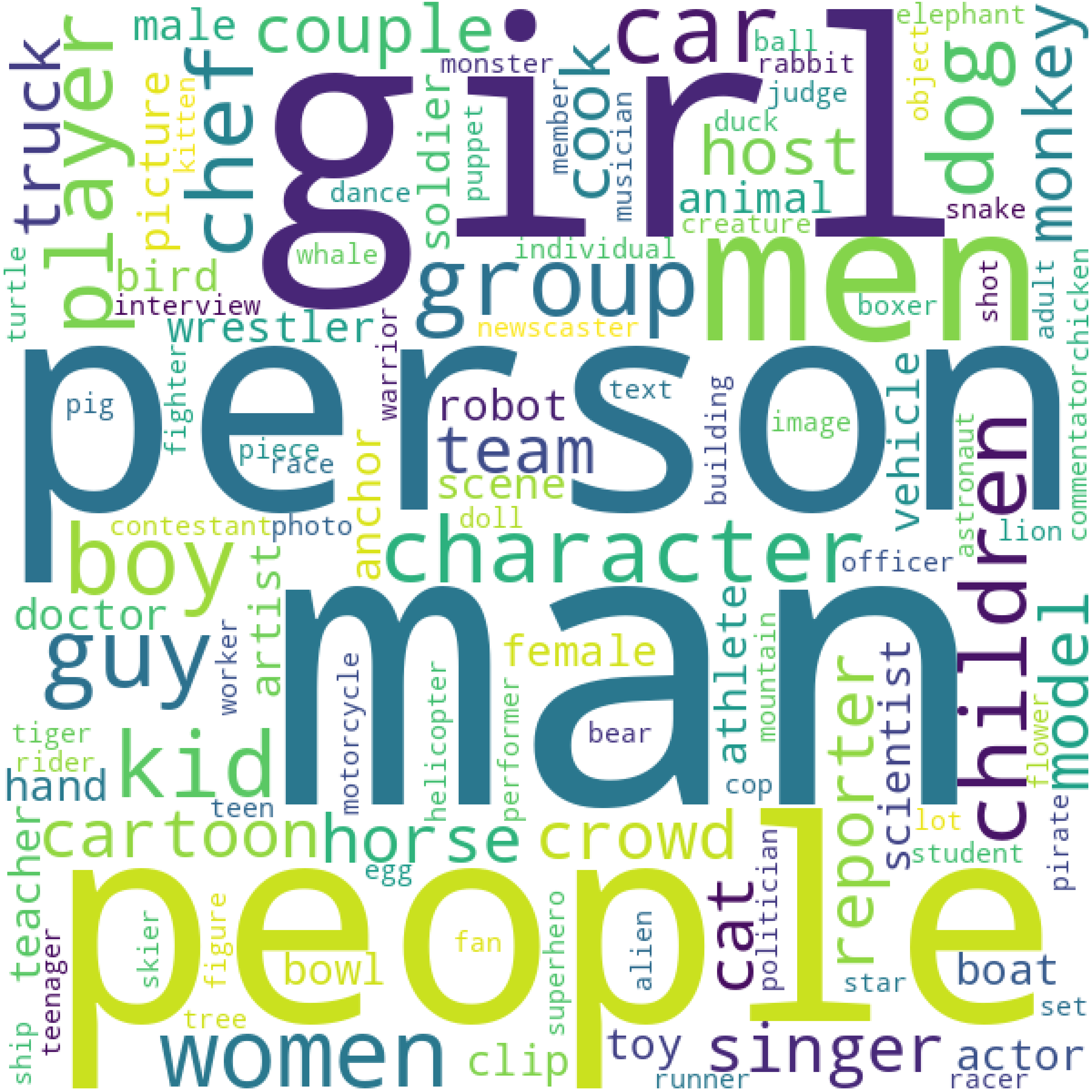}%
    \label{fig:word_cloud_b}%
  }
  \caption{\footnotesize Word cloud visualization with font size proportional to term frequency}
  \label{fig_sim}
\end{figure}

\noindent\textbf{Subject-Oriented MSRVTT}
We apply the same annotation strategy to the
 newly built training, validation, and test splits contain 15,306, 1,085, and 6,979 subject samples,  respectively,  from 6,310, 487, and 2,927  videos, respectively.
Figure~\ref{fig:average_captions_histogram} illustrates the range of subject numbers in each video of subject-oriented MSRVTT, varying from 1 to 10, with the corresponding average number of captions approximately between 1 and 15.
On this dataset, a total of 35,499 regions have been annotated across 25,988 frames. 
Figure~\ref{fig_sim}(b) presents the relative amount of different subjects in the reconstructed MSRVTT dataset.
It shows that the top-4 subjects are slightly different from those of the new MSVD, where  `people' substitutes `cat'.  We also observe that `men', `chef', `car', `player', `guy', `women', `children', and `character' have a similar amount, just next to the top-4 subjects.  

\begin{figure*}[!t]
\centering
\includegraphics[width=0.9\linewidth]{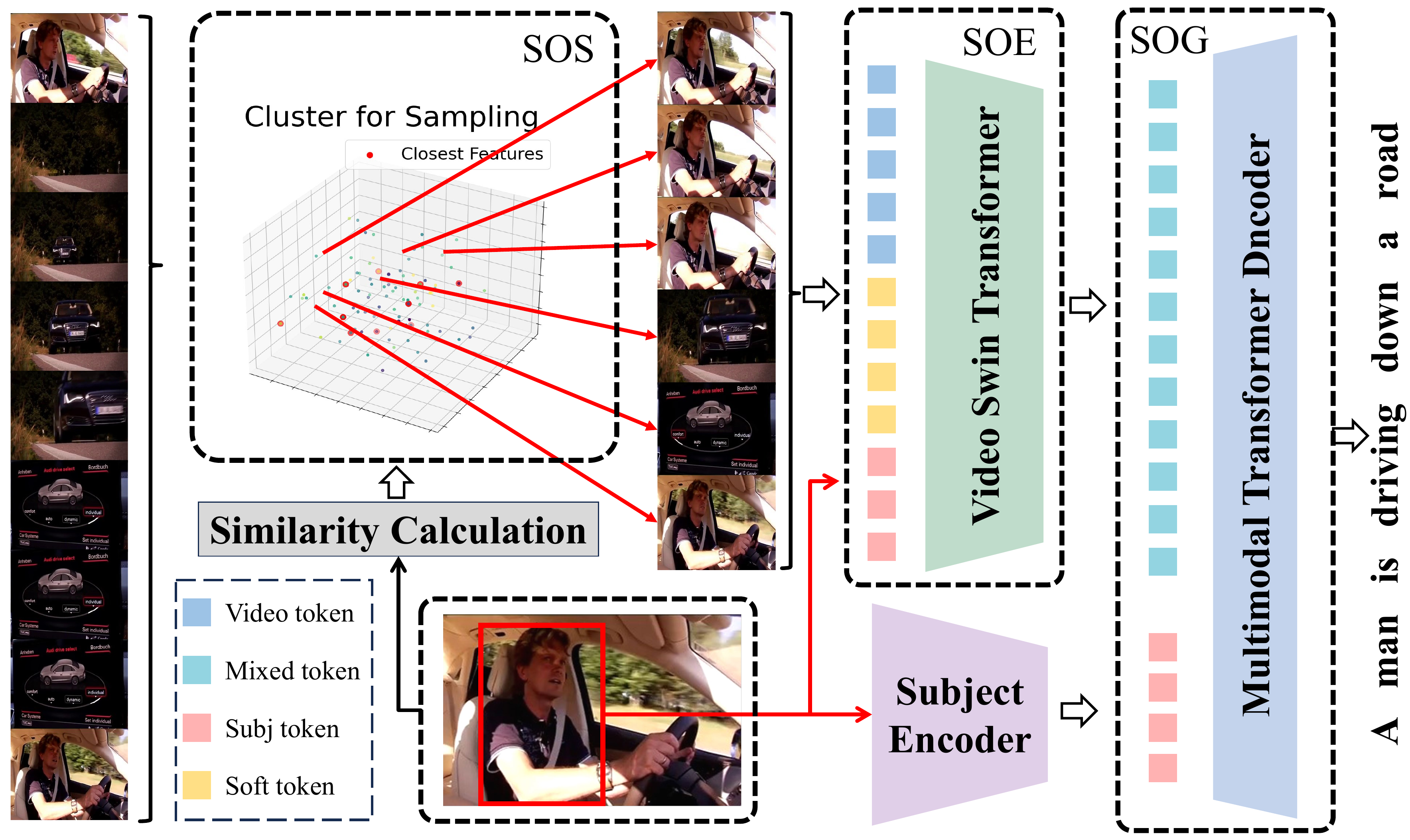}
\caption{An overview of our framework. 
For the video input, we first perform subject-oriented sampling (SOS) to sample the frames related to the subject. Then, we leverage the subject-oriented encoding (SOE) to enhance the model's focus on the subject's activities, tailoring it more effectively to the generation task. Finally, the encoded tokens and subject tokens are concatenated and then fed into the caption generator for description generation.
}
\label{fig:model}
\end{figure*}

\section{Method}
In this section, we first formalize our proposed SOVC task (Sec.~\ref{Task definition}). We then simply extend the existing methods to our task as the baseline (Sec.~\ref{Direct improvement}). Finally, we introduce our proposed SOVCNet, which mainly contains the subject-oriented sampling module and subject-oriented encoding module (Sec.~\ref{sovcnet})

\subsection{Task definition}~\label{Task definition}
Given a test video $\mathbf{V}=[\mathbf{I}_{1},\ldots,\mathbf{I}_{M}]$ of $M$ frames, traditional video captioning methods uniformly sample fixed $N$ frames to represent the whole video and learn a model to describe it. 
Differently, in this paper, our purpose is to select areas of interest in images as subjects and to learn a new model to describe their activities within the video.
\subsection{Simplifying Extension of Existing Methods for Our Task}~\label{Direct improvement}
The workflow of traditional video captioning methods can be described as follows:
\begin{equation}
    \mathbf{C_{\mathbf{V}}} = G_{lan.}(E_{\mathbf{V}}(\mathbf{V}))
\end{equation}
where $E_{\mathbf{V}}$ and $G_{lan.}$ represent the video encoder and language generator,  $\mathbf{C_{\mathbf{V}}}$ is the generated caption about $\mathbf{V}$.
To adapt existing methods to our task, a straightforward idea is to merge subject features with video features, using this combination as the input for the decoder to generate sentences related to the subject. The entire process is illustrated as follows:
\begin{equation}
    \mathbf{C_{\mathbf{S}}} = G_{lan.}(E_{\mathbf{V}}(\mathbf{V}), E_{\mathbf{S}}(\mathbf{S}))
\end{equation}
where $\mathbf{S}$ and $E_{\mathbf{S}}$ are subject area and subject encoder, and $\mathbf{C_{\mathbf{S}}}$ is the generated caption describing the activities of subject. 

However, this approach faces two main challenges: 1) the original uniform video frame sampling method tends to incorporate information irrelevant to the subject, and 2) the modeling of relationships between the subject and video frames is overly direct, making it difficult to adapt to the downstream generation task.
Therefore, we use this straightforward improvement as a simple baseline and introduce the modules for our further enhancements.
\subsection{SOVCNet}~\label{sovcnet}
The workflow of our proposed SOVCNet is shown in Figure~\ref{fig:model}, which first samples video frames related to the selected subject (Section.~\ref{sos}), then encodes the input obtained from the previous step (Section.~\ref{soe}), finally generates sentences that describe the activities of the subject (Section.~\ref{sog}).
\subsubsection{Subject-Oriented Sampling}~\label{sos}
Conventional sampling methods capture one frame at regular intervals, such as every 8 or 32 frames.
However, in the SOVC task, frames that contain information correlated to the subject need to be exploited. To preserve diverse and representative subject-related information, we design a subject-oriented sampling strategy, which contains two steps, i.e. clustering and sampling. 


Specifically, we first extract features of frames using Resnet50 because of its high efficiency. Let $f_{\text{subject}}$ and $D = \{f_1, f_2, \ldots, f_N\}$ denote the features of the subject frame and the set of all frames in the video, respectively.   
Then, we adopt the K-means algorithm to cluster $D$ into $T$ clusters $C = \{C_1, C_2, \ldots, C_T\}$.

To ensure the diversity of sampled video frames, we consider subject-related frames in all clusters. 
For each frame, its cosine similarity to the subject frame is calculated as,
\begin{equation}
s(f_i) = \text{sim}(f_i, f_{subject}), i=1,2,\cdots,N
\end{equation}

Convert these similarities to probabilities using the softmax function:
\begin{equation}
p(f_i \mid C_t) = \frac{e^{s(f_i)}}{\sum_{f_k \in C_t} e^{s(f_k)}}, \quad \forall f_i \in C_t, \text{ for each } C_t \in C
\end{equation}
Then, from each cluster, We sample one frame based on probability.
For simplicity, we use $F \in \mathbb{R}^{T \times H \times W \times C}$ to represent the $T$ sampled frames. To preserve temporal coherence, we maintain the original frame order during sampling, ensuring that the selected frames retain their inherent temporal sequence information.

\subsubsection{Subject-Oriented Encoding}~\label{soe} 
\begin{figure}[!t]
\centering
\includegraphics[width=0.9\linewidth]{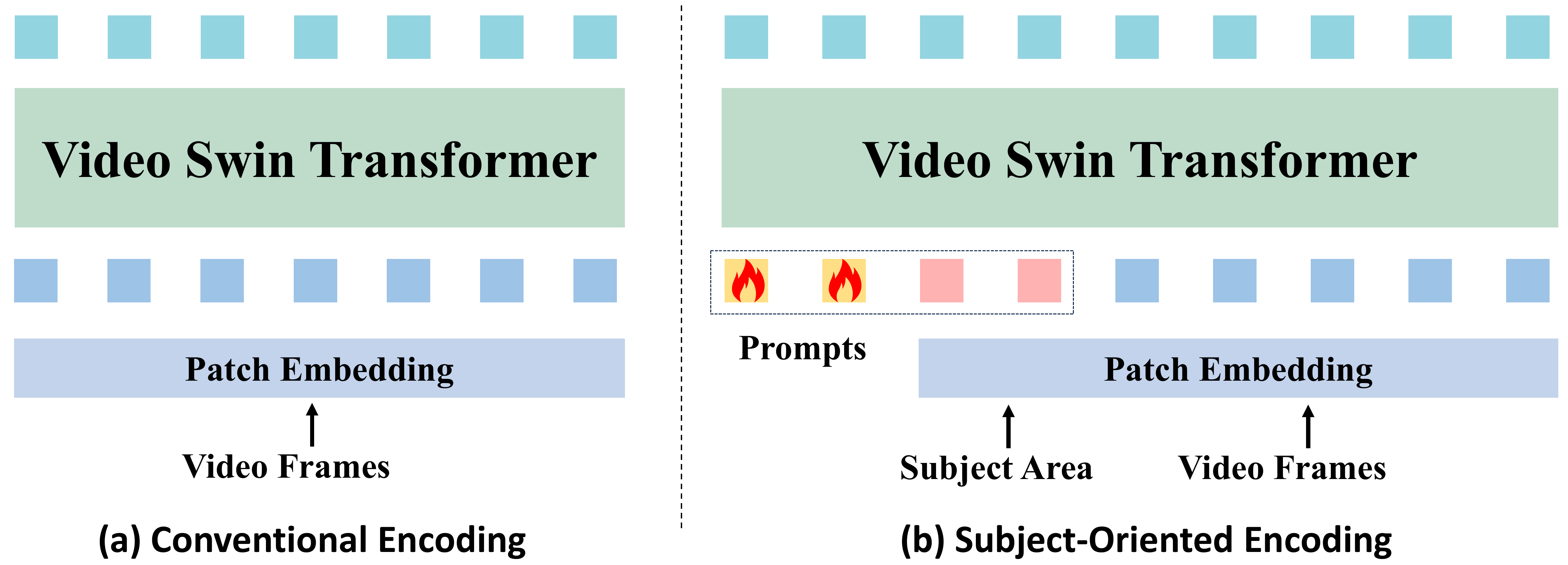}
\caption{Comparison between the Conventional Encoding (a) and Our Subject-Oriented Encoding (b). }
\label{fig:encoding}
\end{figure}

In the encoding phase, we propose a subject-oriented encoding module to make the model focus more on the subject’s activities and adapt to the generation task.
As shown in Figure~\ref{fig:encoding}~(a), the conventional encoding module directly uses the patch embedding operation to process the video frames into a sequence of tokens.
In contrast, we design a subject-oriented encoding module to enable the video encoder to understand the subject we are focusing on.
Specifically, as shown in Figure~\ref{fig:encoding}~(b), 
in addition to performing patch embedding on video frames to obtain a series of tokens, we apply the same operation to the subject area and use it as a visual hard prompt to guide the model's focus on the subject's activities. Meanwhile, we add some learnable tokens as soft prompts to better adapt the model for downstream generative tasks.
This design is grounded in two principles: (1) Visual hard prompts explicitly preserve spatial semantics of user-specified subjects, avoiding ambiguity in object reference (inspired by Visual Prompt Tuning~\cite{jia2022visual}); (2) Learnable soft prompts act as dynamic context adapters, bridging the frozen visual encoder and generative decoder through token-level interaction.
Then, we feed the joint representation of these tokens into the video encoder to obtain the more appropriate feature representation.
The entire process can be described as follows:
\begin{equation}
    \bar{\mathbf{V}} = E_{\mathbf{V}}(Concat(PE(\mathbf{V},\mathbf{S}),\mathbf{P_{soft}}))
\end{equation}
where $PE$ denotes the patch embedding operation, $\mathbf{P_{soft}}$ is the learnable soft prompt, $\bar{\mathbf{V}}$ is the encoded video feature representation.

By combining SOS and SOE modules, our model is capable of fully perceiving the information in the video that is most relevant to the subject, thereby generating more accurate descriptions of the subject.
\subsubsection{Subject-Oriented Caption Generation}~\label{sog}
To generate the subject-oriented caption, we incorporate additional subject information into the caption generation module.
As shown in the right part of Figure~\ref{fig:model}, we use a pre-trained feature extractor and a linear layer as the subject encoder to extract features of the selected subject area and adjust dimensions.
Then, the subject features and video features are concatenated as the input of the language generator to generate the subject-oriented caption $C_{\mathbf{S}}$:
\begin{equation}
    C_{\mathbf{S}} = G_{lan.}(Concat(\bar{\mathbf{V}},E_{\mathbf{S}}(\mathbf{S})))
\end{equation}
This step ensures that the generated caption are aligned with the user-specific subjects in the video.

\section{Experiments}

\subsection{Experimental settings}
\subsubsection{Metrics}
We employ four widely used metrics to assess the quality of the generated captions: BLEU@4 (B@4)~\cite{bleu}, METEOR (M)~\cite{meteor}, ROUGE-L (R)~\cite{rouge}, and CIDEr (C)~\cite{cider}.
Bleu@4 focuses on how closely a translation matches the reference text, Rouge measures the completeness of the content in the generated text, Meteor values adaptability in different expressions, and CIDEr evaluates how the text relates to images.
Thus, CIDEr serves as our primary evaluation metric.

\subsubsection{Baseline models}~\label{baseline models}
As described in Sec.~\ref{Direct improvement}, we directly extend four existing state-of-the-art methods to our task. The descriptions of the original four methods are as follows:

\noindent\textbf{AR-B}~\cite{ARB} belongs to traditional autoregressive captioning models, which use offline ResNet-101 image features and C3D motion features as inputs and employ a transformer decoder as a text generator.

\noindent\textbf{GL-RG}~\cite{DBLP:conf/ijcai/gl-rg} proposes a precisely formulated global-local encoder to exploit rich temporal representation for video captioning.

\noindent\textbf{HMN}~\cite{hmn} proposes a hierarchical modular network that serves as a strong video encoder, which reduces the gap between video and language.
 
\noindent\textbf{SwinBERT}~\cite{swinbert} is the first end-to-end fully Transformer-based model for video captioning. Different from previous methods leveraging off-the-shelf 2D appearance features and 3D motion features, it adopts a trainable VidSwin as a feature extractor to encode spatial-temporal representations. 


\subsubsection{Implementation Details}


For the aforementioned four fundamental video captioning methods, we employ CLIP or Faster R-CNN as the subject encoder.
Among these, we further extend the highest-performing SwinBERT.

In the subject-oriented sampling module, the number of clusters is $32$, which is consistent with the number of sampled frames in SwinBERT. In the subject-oriented encoding module, we set $5$ learnable prompt tokens, whose selection is discussed in Section~\ref{ablation study} (see Fig~\ref{fig:tokens}).

The batch size and learning rate are set to $16$ and $7.5 \times 10^{-5}$ on SO-MSRVTT and SO-MSVD.
We keep other parameters consistent with the original experimental settings of SwinBERT.
The experiments are conducted on multiple Nvidia RTX 3090Ti GPUs and A6000 GPUs.

\begin{table}[!tbp]
  \caption{Results on our Subject-Oriented MSVD and MSR-VTT datasets. Rows with $^{\star}$ denote results on original MSVD and MSRVTT datasets. The best results are highlighted in bold.}
  \centering
  \renewcommand{\arraystretch}{1.3} 
  \setlength{\tabcolsep}{1.5mm} 
  \begin{tabular}{c|c|cccc|cccc}
    \toprule
    \multirow{2}[2]{*}{Models} & \multirow{2}[2]{*}{Year} & \multicolumn{4}{c|}{SO-MSVD} & \multicolumn{4}{c}{SO-MSRVTT} \\
          &      & \multicolumn{1}{c}{B@4} & M & R & C & B@4 & M & R & C \\
    \midrule
    AR-B~\cite{ARB}\_ori$^{\star}$  & 2021 & 48.7  & 33.3    &  -   & 91.8   & 40.5   & 28.7   & -   & 49.1  \\
     AR-B   & 2025 & 27.4  & 22.4     &  50.5    & 69.7   & 17.0   &  16.6  & 38.7   & 43.3 \\
     AR-B\_ext  & 2025  & 31.6  & 24.4  & 53.2   & 78.2   & 20.6    &  18.5  & 42.9   & 55.5 \\
    \hline
    HMN~\cite{hmn}\_ori$^{\star}$ & 2022 & 59.2 & 37.7     & 75.1     &  104.0    &  43.5    &  29.0   &  62.7    & 51.5  \\
    HMN    & 2025 & 29.8  & 22.9   & 51.2   &  77.7   &  17.1    & 16.5   &  39.1    & 44.9 \\
     HMN\_ext  & 2025 & 34.3  & 24.8   & 54.3   &  88.1   &  20.1    & 17.9   &  42.8    & 53.1 \\
    \hline
    GL-RG~\cite{DBLP:conf/ijcai/gl-rg}\_ori$^{\star}$  & 2022 & \multicolumn{1}{c}{45.5} & 30.1  &62.6  &94.3  &55.5  &37.8  &74.7  &51.2 \\
    GL-RG  & 2025 & \multicolumn{1}{c}{34.5} & 26.1 & 54.4 & 86.7 &18.3  & 17.3 & 39.1  & 43.8 \\
    GL-RG\_ext   & 2025  & \multicolumn{1}{c}{36.2} & 27.8 & 56.9  & 96.6  &22.1  &18.8  &43.7   &56.4 \\ 
     \hline
     SwinBert~\cite{swinbert}\_ori$^{\star}$ & 2022 & \multicolumn{1}{c}{55.7} &39.7  &75.7  &109.4  & 41.9 & 29.9 & 62.1  & 53.8 \\
    SwinBert   & 2025 & \multicolumn{1}{c}{33.4} & 25.7 &54.2  &89.6  &18.2  & 17.4 & 40.4  & 46.9 \\
    SwinBert\_ext   & 2025  & \multicolumn{1}{c}{ {35.8}} &  27.1  & 56.3 & 97.7  &  21.5 & 18.9  & 44.3 & 57.1 \\ 
     \hline
     SOVCNet (Ours)   & 2025  & \multicolumn{1}{c}{ \textbf{36.5}} & \textbf{ 27.6}   & \textbf{57.6}  &\textbf{98.8}   & \textbf{ 22.1}  &\textbf{19.1}   & \textbf{45.5}  & \textbf{61.1} \\ 
    \bottomrule
    \end{tabular}
  \label{compare}
\end{table}

\begin{table}[h]
  \caption{Subject accuracy of different methods on our Subject-Oriented MSVD and MSR-VTT benchmarks.}
  \centering
  \begin{tabular}{l|cc}
    \toprule
    \multirow{2}[2]{*}{Methods}      & SO-MSVD     & SO-MSRVTT \\
               & Acc & Acc      \\
    \midrule
    AR-B & 52.5\% & 46.5\% \\
    AR-B\_ext & 59.1\% & 59.9\% \\
    \midrule
    
    HMN & 59.6\% & 47.4\% \\
    HMN\_ext & 65.2\% & 60.8\% \\
    \midrule
    GL-RG & 56.5\% & 45.6\% \\
    GL-RG\_ext & 60.0\% & 57.4\% \\
    \midrule
    SwinBERT & 60.9\% & 48.8\% \\
    SwinBERT\_ext & 63.9\% & 61.0\% \\
    \midrule
    SOVCNet & \textbf{65.7\%} & \textbf{63.7\%} \\
    \bottomrule
  \end{tabular}
  \label{Acc}
\end{table}

\subsection{Subject-Oriented Video Captioning Performance}
To the best of our knowledge, no methods are doing precisely the same task as ours before.
Therefore, we use the four extended methods mentioned in Section~\ref{baseline models} for comparison.
Table~\ref{compare} shows the performance comparison between our method and other baseline models on the SO-MSVD and SO-MSRVTT test sets. 
For each extended baseline model, the first and second rows present their performance on the origin datasets and our reproduced datasets, respectively.
The results show that: (I) The four state-of-the-art methods encounter a large performance drop on our new task compared to the performance on the original video captioning task. For example, on our subject-oriented MSVD, the CIDEr score of AR-B decreases by 31.7\% relative to itself, 
while HMN reduces by 33.8\% compared to its own performance.
These indicate our subject-oriented video captioning task poses new challenges to existing methods. To succeed, they should have the ability to be aware of the specified target, and the ability to distinguish related activities.
(II) Our extension of the four state-of-the-art methods brings large improvement. For example, on subject-oriented MSRVTT, our extended SwinBERT's CIDEr score improves by 21.3\% compared to without extension, while the extension of GL-RG exhibits an increase of 28.8\%.
These demonstrate the effectiveness of our utilization of user-specified targets. As these are intuitive extensions, we believe future dedicatedly designed methods could further boost the caption quality.
(III) Our SOVCNet achieves the best performance across all metrics compared to the other four extended methods. Specifically, compared to the second-best SwinBert\textunderscore ext, our SOVCNet gains 4.0 and 1.1 absolute improvements in terms of the main metric CIDEr on SO-MSRVTT and SO-MSVD, respectively.
Meanwhile, we observe 0.6 and 0.7 absolute improvements in terms of another main metric BLEU@4. The large improvement demonstrates the effectiveness of our proposed SOVCNet.

In addition to using four conventional video description quality assessment metrics, we also compare the subject accuracy of SOVCNet against the original methods on the test set of the new datasets. 
If the predicted subject is found within the set of ground truth subjects, the prediction is considered correct; otherwise, it is deemed incorrect.

As shown in Table~\ref{Acc}, the accuracy of the four extended methods has also further improved compared to the original four methods.
Additionally, our SOVCNet achieves the highest accuracy of 65.7\% and 63.7\% on SO-MSVD and SO-MSRVTT, leading by 1.8\% and 2.7\% absolute improvements over the second-best SwinBERT\_ext.
This shows that SOVCNet not only produces higher-quality captions but also more accurately predicts the subject specified by the user.

\begin{table}
  \caption{Performance comparison of different modules in SOVCNet on SO-MSVD and SO-MSRVTT. The best results are highlighted in bold.}
  \centering
  \renewcommand{\arraystretch}{1.3} 
  \setlength{\tabcolsep}{1.5mm} 
  \begin{tabular}{c|cccc|cccc}
    \toprule
    \multirow{2}[2]{*}{Model} & \multicolumn{4}{c|}{SO-MSVD} & \multicolumn{4}{c}{SO-MSRVTT} \\
          & B@4 & M & R & C & B@4 & M & R & C \\
    \midrule
    SwinBERT\_ext (Baseline) & 35.8 & 27.1 & 56.3 & 97.7 & 21.5 & 18.9 & 44.3 & 57.1 \\
    SwinBERT\_ext + SOS & 36.5 & 27.1 & 56.4 & 98.3 & 21.7 & 19.1 & 44.4 & 59.2 \\
    SwinBERT\_ext + SOE & 35.2 & 26.9 & 57.5 & 96.6 & 20.6 & \textbf{19.2} & 43.4 & 51.9 \\
    SOVCNet (Ours) & \textbf{36.5} & \textbf{27.6} & \textbf{57.6} & \textbf{98.8} & \textbf{22.1} & {19.1} & \textbf{45.5} & \textbf{61.1} \\
    \bottomrule
  \end{tabular}
  \label{proposed module}
\end{table}

\begin{table}[htbp]
  \caption{Comparison of different sampling strategies on SO-MSRVTT.}
  \label{tab:balanced_table}
  \centering
  \small
  \setlength{\tabcolsep}{5pt}
  \renewcommand{\arraystretch}{1.1}
  \begin{tabular}{@{}l !{\vrule width 0.8pt} *{4}{w{c}{1.6cm}} @{}} 
    \toprule
    \textbf{Strategy} & \textbf{B@4} & \textbf{M} & \textbf{R} & \textbf{C} \\
    \midrule
    Regular          & 20.6  & 19.2  & 43.4  & 51.9  \\
    Similarity       & 21.5  & 19.2  & 44.8  & 58.1  \\
    Adding-interval  & 21.9  & \textbf{19.3} & 45.0  & 59.2  \\
    Clustering       & \textbf{22.1} & 19.1  & \textbf{45.5} & \textbf{61.1} \\
    \bottomrule
  \end{tabular}
\end{table}

\subsection{Ablation Study}\label{ablation study}
In this section, we present the ablation study results of our SOVCNet. We first ablate the effectiveness of our proposed SOS and SOE modules.
Then, we demonstrate the impact of different SOS strategies and the number of SOE soft tokens. Finally, we show the effect of using different subject features in the generation phase.

\noindent\textbf{Effectiveness of Proposed Modules.}
To evaluate the effectiveness of our proposed key modules on overall performance, we do ablation experiments on both SO-MSVD and SO-MSRVTT.
The results are presented in Table~\ref{proposed module}.
The SOS module makes the greatest contribution, increasing the CIDEr score to 59.2 from 57.1 on SO-MSRVTT, and to 98.3 from 97.7 on SO-MSVD.
Our full model achieves the best performance, with the CIDEr score reaching 61.1 and 98.8 on SO-MSRVTT and SO-MSVD.
However, performance drops when solely using SOE. This may be because we adjust the model's frame encoding process without altering the frame sampling, resulting in a reduced ability to encode standard frames.

\noindent\textbf{Effect of different sampling strategies.}
To study the effect of different sampling strategies on our method, we compare the performance of four different sampling strategies on SO-MSRVTT.
The first row represents regular uniform sampling.
The second row converts the similarity between all frames and the subject frames into probabilities for probabilistic sampling.
The third row adds intervals on the basis of the second row to increase the diversity of information.
The last row clusters video frames into k categories, and then probabilistically selects one frame from each category.
%
The comparison results show that our cluster-based sampling method achieves the highest CIDEr score of 61.1 on SO-MSRVTT, exceeding other methods by 1.9 to 9.2.

\begin{figure}[!h]
\centering
\includegraphics[width=0.8\linewidth]{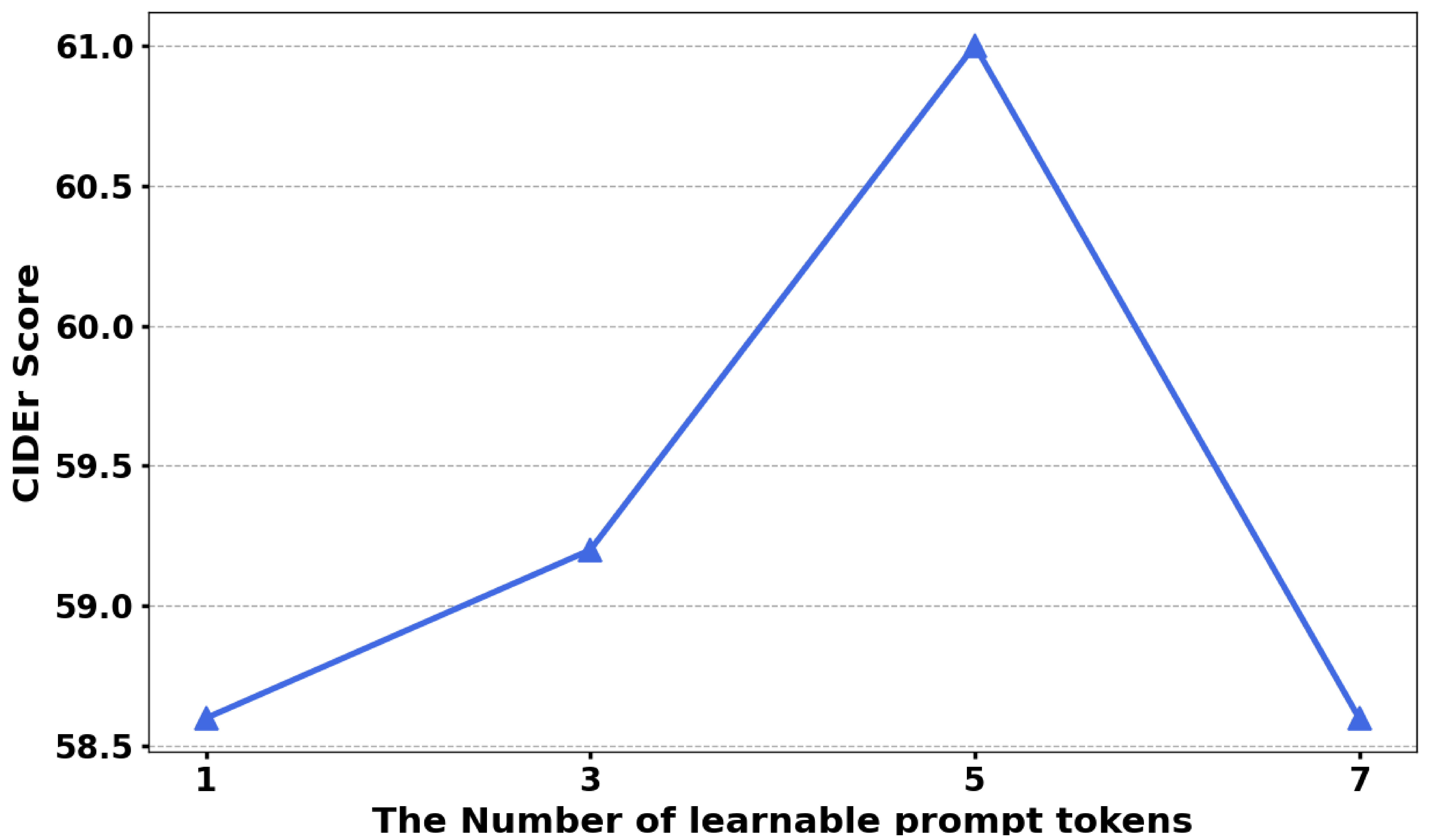}
\caption{Comparison of number of Soft tokens on SO-MSRVTT.}
\label{fig:tokens}
\end{figure}

\begin{table}[!htbp]
  \centering
  \caption{Results under different visual subject encoders.}
  \renewcommand{\arraystretch}{1.3} 
  \setlength{\tabcolsep}{1.5mm} 
  
  \resizebox{\columnwidth}{!}{ 
    \begin{tabular}{c|cccc|cccc}
      \toprule
      \multirow{2}[2]{*}{Method} & \multicolumn{4}{c|}{SO-MSVD} & \multicolumn{4}{c}{SO-MSRVTT} \\
      & B@4 & M & R & C & B@4 & M & R & C \\
      \midrule
        AR-B\_ext~(Faster-RCNN)  &30.3   &23.8   &52.5    & 76.2   & 19.3    & 17.6   & 41.8   & 50.8 \\
        AR-B\_ext~(CLIP V/L-14)  & 31.6  & 24.4  & 53.2   & 78.2   & 20.6    &  18.5  & 42.9   & 55.5  \\
    \hline
        HMN\_ext~(Faster-RCNN)   & 32.7  & 24.1  & 53.6   & 85.2    & 19.8     & 17.6   & 42.5     & 52.2 \\
        HMN\_ext~(CLIP V/L-14)   & 34.3  &24.8   & 54.3   &  88.1   &  20.1    & 17.9   &  42.8    & 53.1 \\
    \hline
        GL-RG\_ext~(Faster-RCNN)     & \multicolumn{1}{c}{37.8} &27.5  & 56.5  &95.9   &20.8      &18.2  &42.0   & 53.3  \\ 
        GL-RG\_ext~(CLIP V/L-14)     & \multicolumn{1}{c}{36.2} & 27.8 & 56.9  & 96.6  &22.1  &18.8  &43.7   &56.4   \\ 
     \hline
      SwinBERT\_ext~(Faster-RCNN)   & 35.6 & 26.7 & 56.2 & 96.8 & 21.2 & 19.1 & 44.1 & 56.8 \\
      SwinBERT\_ext~(CLIP V/L-14)   & 35.8 & 27.1 & 56.3 & 97.7 & 21.5 & 18.9 & 44.3 & 57.1 \\
    \hline
        SOVCNet~(Faster-RCNN) & {35.9} & {27.1} & {56.6} & {97.8} & {21.4} & {19.1} & {44.5} &{58.1} \\
       SOVCNet~(CLIP V/L-14) & \textbf{36.5} & \textbf{27.6} & \textbf{57.6} & \textbf{98.8} & \textbf{22.1} & \textbf{19.1} & \textbf{45.5} & \textbf{61.1} \\
      \bottomrule
    \end{tabular}
}
  \label{use different subject features}
\end{table}

\noindent\textbf{Effect of the number of soft tokens.} We explore the effect of the number of soft tokens, and Figure~\ref{fig:tokens} shows that 5 is the better choice than other numbers.
In Figure~\ref{fig:tokens}, the CIDEr score initially drops as the number increases from 1 to 3, then reaches its highest point from 3 to 5, and subsequently falls again from 5 to 7.
Consequently, we set 5 as the standard number of soft tokens.

\noindent\textbf{Effect of different subject features.}
In Table~\ref{use different subject features}, we compare the performance of our method using Faster-RCNN and VIT(L-14)-based CLIP to extract subject features, respectively.
It shows that CLIP V/L-14 leads to better results for all methods on the SO-MSVD dataset in terms of the main metric CIDEr. The improvements range from 0.9 to 2.9. 
And the performance of the SO-MSRVTT dataset also significantly improves. 
For example, SOVCNet increases the CIDEr score from 58.1 to 61.1 on SO-MSRVTT.
Overall, using CLIP-VIT features achieves better performance.

\subsection{Qualitative Results}
\begin{figure*}[!ht]
\centering
\includegraphics[width=\linewidth]{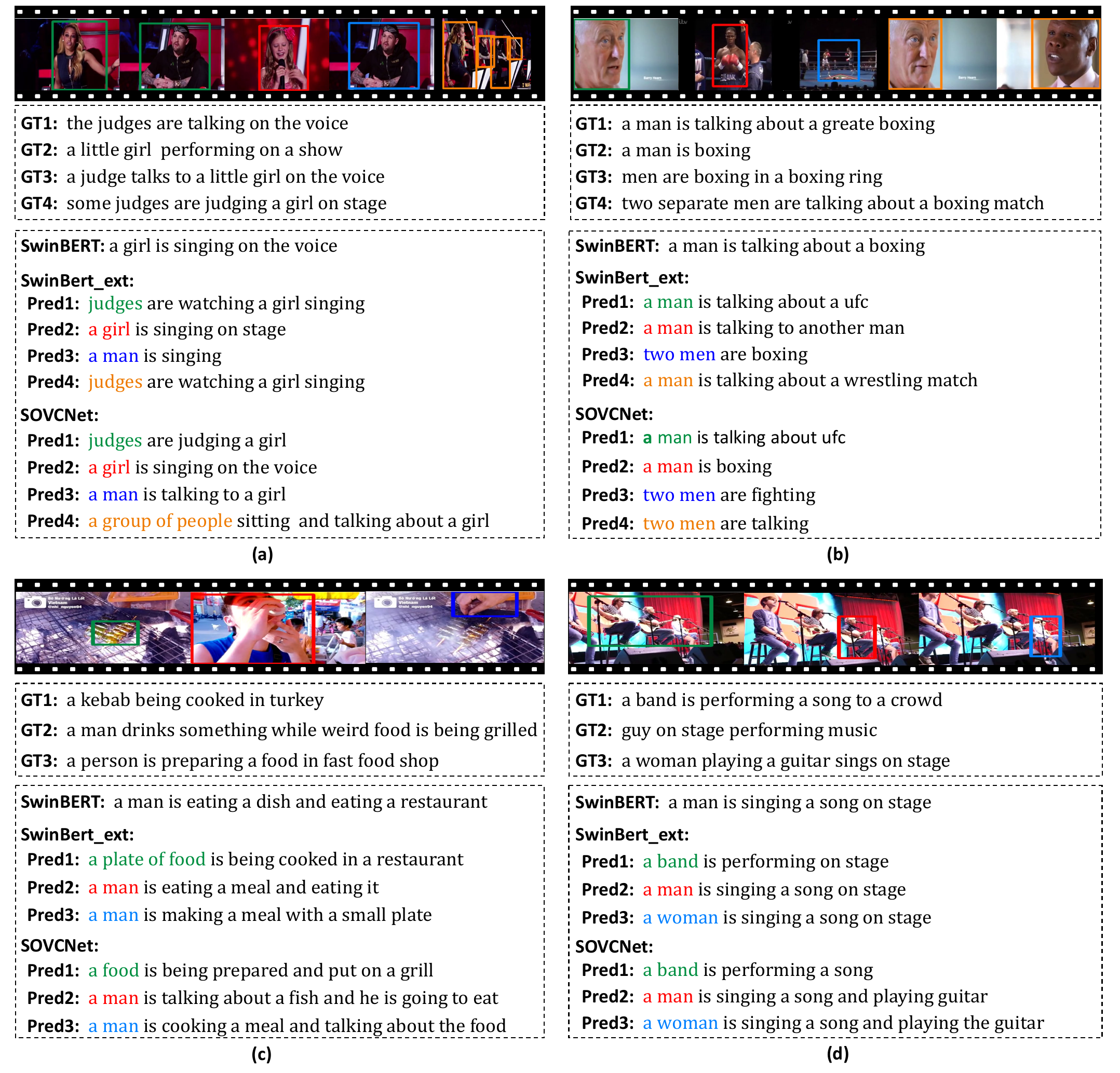}
\caption{Qualitative results on Subject-Oriented MSR-VTT.}
\label{fig:result}
\end{figure*}
We show some qualitative results in Figure~\ref{fig:result}.
Compared to the origin SwinBERT and extended SwinBERT, our SOVCNet better focuses on and accurately describes the entities of interest.
For instance, the video in Figure~\ref{fig:result} (a) shows a voice show. It can be observed that the origin SwinBERT only describes ``a girl''. Compared to the extended SwinBERT, our method generates accurate descriptions for the girl, the judges, and one of the judges, as well as their actions.
Figure~\ref{fig:result} (b) displays two men talking about a boxing match. 
The extended SwinBERT incorrectly identified two speakers in a conversation as ``a man is talking about a wrestling match'',  while SOVCNet accurately describes it as ``two men are talking''.
In Figure~\ref{fig:result} (c), the extended SwinBERT only roughly describes ``a plate of food is being cooked in a restaurant''. In contrast, our improved method captures ``a grill'' in more detail.
In Figure~\ref{fig:result} (d), compared to Pred2 ``a man is singing a song on stage'' in the extended SwinBERT, SOVCNet is capable of describing the ``playing guitar'' action appearing in the video.
The qualitative results further prove the effectiveness of our method and the significance of this task.
\vspace{-2mm}

\section{Conclusion}
In this paper, we propose a novel subject-oriented video captioning task that requires the model to generate subject-oriented video descriptions based on the subject specified by the user. To solve this task, we propose a corresponding method called SOVCNet. This approach begins by sampling frames related to the subject to minimize irrelevant information. Then it uses the subject areas as hard prompts and plugs the soft prompts to enhance the model's focus on the subject's activities, aiding adaptation to the generative task. Additionally, we incorporate extra subject information to help the model generate accurate descriptions.
To evaluate our system, we construct two new datasets using the public MSVD and MSR-VTT datasets. Qualitative and quantitative experimental results demonstrate the effectiveness of our method.

\bibliographystyle{elsarticle-num}
\bibliography{editable}

\end{document}